\documentclass[conference]{IEEEtran}
\usepackage{caption}

\usepackage{amsthm,amsmath,amssymb}
\usepackage[table]{xcolor}  % For cell colors
\usepackage{todonotes}
\usepackage[T1]{fontenc}
\usepackage{cite}
\usepackage{svg}
\usepackage{amsfonts}
\usepackage{todonotes}
\usepackage{graphicx}
\usepackage{textcomp}
\usepackage{xcolor}
\usepackage{orcidlink}
\def\BibTeX{{\rm B\kern-.05em{\sc i\kern-.025em b}\kern-.08em
    T\kern-.1667em\lower.7ex\hbox{E}\kern-.125emX}}
\usepackage{changepage}  % For adjustwidth environment
\usepackage{threeparttable}
\usepackage{wrapfig}
\usepackage{array}  % For better column alignment
\usepackage[noend]{algpseudocode}
\usepackage{algorithm}%algorithmic
\usepackage{adjustbox}
\usepackage{amsmath}
\usepackage{tabularx}
\usepackage{booktabs}
\usepackage{makecell}
\usepackage{multirow} % Required for multirows
\usepackage{bbding}
\hypersetup{hidelinks,
	colorlinks=true,
	allcolors=black,
	pdfstartview=Fit,
	breaklinks=true}
\usepackage[marginal]{footmisc}

\DeclareMathOperator*{\argmin}{arg\,min}
\begin{document}
%
% \title{Contribution Title}
\title{DATTA: Domain Diversity Aware Test-Time Adaptation for Dynamic Domain Shift Data Streams}
\IEEEoverridecommandlockouts
\author{
    \IEEEauthorblockN{%
        Chuyang Ye\textsuperscript{1}\IEEEauthorrefmark{2}, 
        Dongyan Wei\textsuperscript{1}\IEEEauthorrefmark{2}, 
        Zhendong Liu\textsuperscript{1}, 
        Yuanyi Pang\textsuperscript{1}, 
        Yixi Lin\textsuperscript{1}, \\
        Qinting Jiang\textsuperscript{2}, 
        Jingyan Jiang\textsuperscript{1}\IEEEauthorrefmark{1}, 
        Dongbiao He\textsuperscript{3}
    }
    \IEEEauthorblockA{\textsuperscript{1}Shenzhen Technology University}
    \IEEEauthorblockA{\textsuperscript{2}Tsinghua University}
    \IEEEauthorblockA{\textsuperscript{3}Computer Network Information Center, Chinese Academy of Sciences}
    \IEEEauthorblockA{Emails: \{youngyorkye, eastendwy, dong2172go, pyyaiiaii, llinxiyi\}@gmail.com,}
    \IEEEauthorblockA{jqt23@mails.tsinghua.edu.cn, jiangjingyan@sztu.edu.cn, dbhe@cnic.cn}
    \vspace{-0.3in}
    \thanks{~~~This study was supported by the Natural Science Foundation of Top Talent of SZTU (Grant No. GDRC202413).}
    \thanks{~~~\IEEEauthorrefmark{1}Corresponding author. \IEEEauthorrefmark{2}Contributed equally.}
}

\maketitle              % typeset the header of the contribution

\begin{abstract}
\sloppy
% Underfull \hbox (badness 1939) in paragraph at lines 201--205
% TODO The abstract should briefly summarize the contents of the paper in 150--250 words.

Test-Time Adaptation (TTA) addresses domain shifts between training and testing. However, existing methods assume a homogeneous target domain (e.g., single domain) at any given time. They fail to handle the dynamic nature of real-world data, where single-domain and multiple-domain distributions change over time. We identify that performance drops in multiple-domain scenarios are caused by batch normalization errors and gradient conflicts, which hinder adaptation. To solve these challenges, we propose Domain Diversity Adaptive Test-Time Adaptation (DATTA), the first approach to handle TTA under dynamic domain shift data streams. It is guided by a novel domain-diversity score. DATTA has three key components: a domain-diversity discriminator to recognize single- and multiple-domain patterns, domain-diversity adaptive batch normalization to combine source and test-time statistics, and domain-diversity adaptive fine-tuning to resolve gradient conflicts. Extensive experiments show that DATTA significantly outperforms state-of-the-art methods by up to 13\%. Code is available at \href{https://github.com/DYW77/DATTA}{https://github.com/DYW77/DATTA}.

\end{abstract}


% \keywords{Quality of Experience \and Test-time Adaptation \and Test-time Normalization \and Domain Generalization \and Domain Adaptation.}
% 
%
%
%
% \footnote{First Author and Second Author contribute equally to this work.\\}
% \footnote{ 
%     \textbf{*} Equal Contribution\\
%     \textbf{\Envelope} Corresponding Authors
% }
\section{Introduction}

\label{sec:intro}
Deep neural networks (DNNs) have achieved remarkable success, yet their performance often degrades when the distributions of the training and testing domains differ \cite{hendrycks2019robustness,choi2021robustnet,recht2019imagenet}. Such domain shifts frequently occur in real-world applications, due to weather variations or sensor degradation, potentially resulting in catastrophic outcomes in critical fields such as healthcare and autonomous driving.
\begin{figure}
    \centering
    \includegraphics[width=0.95\linewidth]{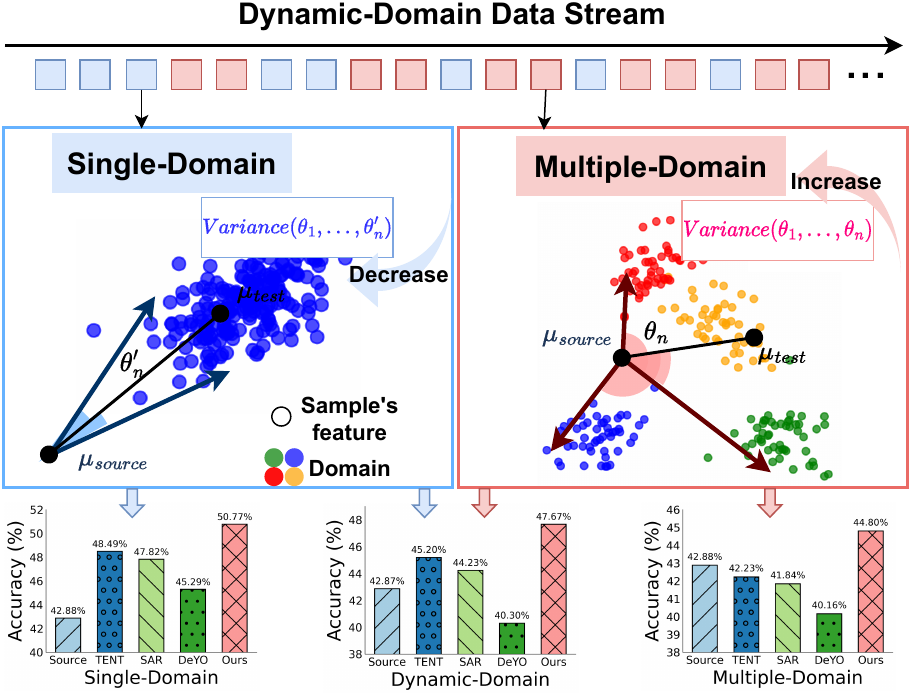}
    % \vspace{-0.2in}
    \caption{The top illustrates the Dynamic Domain Shift Data Streams, while the bottom compares classification Top1-accuracy (\%) on the CIFAR-100-C dataset using EfficientViT-M5, showing the superior performance of our method (DATTA) over previous TTA methods under dynamic domain shift data streams.}
    \label{fig:intro}
    \vspace{-0.25in}
\end{figure}

Recently, numerous test-time adaptation (TTA) methods have been proposed to address domain shifts. TTA enables pre-trained models to adapt without supervision by using the incoming testing data stream for real-time adjustments, without relying on training data and ground truth labels \cite{wangTentFullyTesttime2021b}. Existing TTA methods typically involve two main steps: (1) (Re-)calibrating batch normalization (BN) statistics. For example, NOTE \cite{gong2022note} corrects BN with instance norm. (2) Optimizing model parameters. TENT \cite{wangTentFullyTesttime2021b} optimizes the model during testing to minimize the entropy of its predictions by modulating its features. And DeYO \cite{lee2024entropy} uses pseudo-label probability differences to identify harmful samples and adapt models by prioritizing shape information.%SAR \cite{niu2023towards} selectively minimizes entropy by excluding noisy samples and optimizing model stability. And RoTTA \cite{yuanRobustTestTimeAdaptation2023} simulates an i.i.d. data stream by constructing a sampling pool and adapting BN statistics.
% \cite{niu2023towards,yuanRobustTestTimeAdaptation2023,liu2023vida,lee2024entropy}. 
% \todo{add some ref
 % and illustrate some category of tta...tent...note..rotta....deyo...}

% \textbf{TENT} \cite{wangTentFullyTesttime2021b} minimizes prediction entropy to boost confidence, estimates normalization statistics, and updates channel-wise transformations online.
% \textbf{NOTE} \cite{gongNOTERobustContinual2023} corrects normalization for out-of-distribution samples and simulates an i.i.d. data stream from a non-i.i.d. stream.
% \textbf{SAR} \cite{niu2023towards} selectively minimizes entropy by excluding noisy samples and optimizing stability.
% \textbf{RoTTA} \cite{yuanRobustTestTimeAdaptation2023} simulates an i.i.d. data stream by constructing a sampling pool and adapting BN statistics.
% \textbf{ViDA} \cite{liu2023vida} decomposes features into high- and low-rank components for knowledge sharing.
% \textbf{DeYO} \cite{lee2024entropy} uses pseudo-label probability differences to identify harmful samples and adapt models by prioritizing shape information.

While current TTA models have demonstrated notable success, their effectiveness remains largely constrained to an idealized data stream where test samples are assumed to be ``Static'', meaning that they exhibit consistent types of domain shifts. 
Although recent studies have explored adapting to dynamic target distributions in evolving environments~\cite{wangContinualTestTimeDomain2022}, and SAR specifically addressed data streams involving multiple-domain settings~\cite{niu2023towards}, these approaches typically assume that, at any given time, the data stream only belongs to a single domain or consists of multiple domains. 
% Although recent studies COTTA have explored adapting to dynamic target distributions in evolving environments~\cite{wangContinualTestTimeDomain2022}, and SAR specifically addressed data stream involving multiple-domain settings~\cite{niu2023towards}, these approaches typically assume that, at any given time, the data stream only belongs to a single domain or consists of multiple domains. 
However, such methods fail to effectively capture the ``Dynamic'' inherent in real-world data streams. In reality, domain shifts are often neither gradual nor confined to a single domain. Instead, they tend to be unpredictable, potentially involving multiple domains simultaneously, while reverting to a single domain in the next moment, referred to as \textit{dynamic domain shift data streams}. For instance, in autonomous driving systems, the input data stream can abruptly shift from a single-domain data stream, such as driving on a clear highway during the day, to a multiple-domain data stream, such as navigating through a crowded urban environment at night with rain. Such dynamic and heterogeneous changes pose significant challenges to existing TTA methods.
As shown in Fig.~\ref{fig:intro}, under dynamic domain shifts, all methods experience a performance drop compared to the Single-Domain scenario, with declines of 3.29\% for TENT, 3.59\% for SAR, and 4.99\% for DeYO.
% \todo{results}

% Existing TTA methods largely assume that test data are drawn from a single domain, where test samples within a batch share similar characteristics \cite{wangTentFullyTesttime2021b,wangContinualTestTimeDomain2022,gongNOTERobustContinual2023,yuanRobustTestTimeAdaptation2023}. Recently, Niu et al. \cite{niu2023towards} highlighted that real-world applications often involve batches containing samples from multiple domains. However, neither of these assumptions fully captures the complexity of real-world scenarios and it's intrinsically \textbf{static} because of the stationary experimental settings. Real-world applications are inherently \textbf{dynamic} and feature a mixture of single-domain and multiple-domain patterns. For instance, during autonomous driving in urban areas, a car may encounter sequences of relatively stable scenes (e.g., driving on a straight road with consistent traffic flow and lighting conditions), followed by highly dynamic situations (e.g., entering a busy intersection with varying vehicle types, pedestrians, and lighting changes). These dynamic domain conditions pose unique challenges for TTA, as traditional methods struggle to maintain robustness and adapt effectively in such evolving environments.

\begin{figure*}[t]
    \centering
        \centering
        \includegraphics[width=1\linewidth]{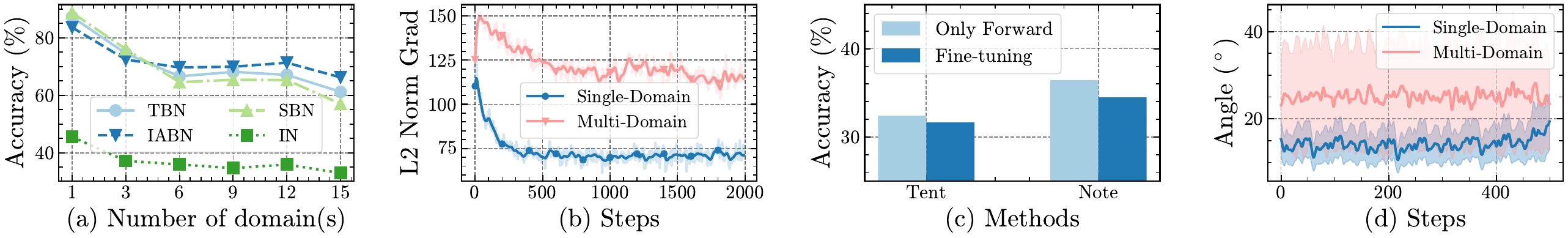}
    % \caption{ResNet-50 experimental analysis: \textbf{(a)} Impact of Domain Number on Accuracy: Illustration of the effect of varying domain numbers on model accuracy using CIFAR-10-C at severity level 5. \textbf{(b)} L2 Norm of Gradient Dynamics: Comparison of the L2 norm of gradient dynamics under two different data patterns using CIFAR-10-C at severity level 5. \textbf{(c)} Impact of methods' Only Forward and Fine-Tuning on accuracy by using CIFAR-100-C at Severity Level 5 and \textbf{(d)} Domain-diversity angle across two data patterns by using CIFAR-100-C at Severity Level 5.}
\vspace{-0.25in}
\caption{\textbf{(a)} Impact of number of domain(s) on Accuracy: Illustration of the effect of varying number of domains on accuracy. \textbf{(b)} L2 Norm of Gradient Dynamics: Comparison of the L2 norm of gradient under two different data patterns. Both (a) and (b) use CIFAR-10-C at severity level 5 and ResNet-50. \textbf{(c)} Impact of Only Forward and Fine-Tuning on accuracy. \textbf{(d)} Domain-Diversity angle across two data patterns. Both (c) and (d) use CIFAR-100-C at severity level 5 and ResNet-50.}
% \caption{ResNet-50 experimental analysis: \textbf{(a, b)} Using CIFAR-10-C at severity level 5: (a) Impact of domain number on accuracy, and (b) L2 norm of gradient dynamics under two different data patterns. \textbf{(c, d)} Using CIFAR-100-C at severity level 5: (c) Impact of "Only Forward" and "Fine-Tuning" methods on accuracy, and (d) Domain-diversity angle across two data patterns.}
    % Degradation of indiscriminate fine-tuning using CIFAR-100-C at severity level 5
\vspace{-0.25in}
    \label{fig:motivation-all}
\end{figure*}

To address this, we conduct an in-depth investigation into the underlying causes of performance degradation under such dynamic domain shifts. 
As analyzed in  \S Sec.~\ref{sec:motivation}, these challenges stem from two core issues under multiple-domain patterns:
(1) Vanilla batch normalization struggles in multiple-domain patterns as it averages statistics across the entire batch, blending data from multiple domains. This leads to inaccurate statistics and distorted feature representations.
% vanilla test-time batch normalization technique in TTA averages over the entire batch, blending the statistics from multiple domains. This mixing results in unrepresentative estimates and mismatched feature normalization. 
(2) Such mismatched feature normalization further causes gradient instability, disrupting back-propagation and hindering effective optimization and convergence.

Recognizing these issues, we argue that the key lies in dynamically identifying the multiple-domain pattern and enhancing the reliability of batch normalization statistics tailored for multiple-domain scenarios. Based on this insight, we propose a robust TTA method under dynamic domain shift data streams, called \textbf{D}omain Diversity \textbf{A}daptive \textbf{T}est-\textbf{T}ime \textbf{A}daptation (DATTA), which introduces an innovative approach to quantify and address domain diversity. The overall framework of DATTA is illustrated in Fig.~\ref{fig:overview}. DATTA introduces a lightweight Domain-Diversity Discriminator (DD), with the Domain-Diversity Score at its core: an innovative metric that dynamically evaluates the alignment between individual samples and batch-level distributions. This score leverages batch normalization statistics and feature maps, providing a principled way to quantify domain diversity in real-time.  To ensure robust adaptation, we further integrate a kernel density estimation based adaptive threshold, which dynamically separates single-domain and multiple-domain patterns by uncovering the latent structure of domain-diversity score distributions.  Building on this score, DATTA employs two adaptive adjustment mechanisms to ensure robust test-time adaptation: (1) Domain-Diversity Adaptive Batch Normalization (DABN), which dynamically aggregates source and test-time statistics for robust feature normalization across diverse domains, and (2) Domain-Diversity Adaptive Fine-Tuning (DAFT), a selective fine-tuning mechanism that prevents harmful updates caused by gradient conflicts in multiple-domain scenarios. Together, these components enable DATTA to achieve significant improvements in accuracy and efficiency, outperforming state-of-the-art methods. Our contributions are summarized as follows:

First, we \textit{firstly} identify the problem of \emph{Dynamic Domain Shifts} in TTA and analyze its underlying challenges. Our motivation experiments reveal that the performance degradation in current TTA methods arises from the failure of batch normalization and conflicts in gradient optimization.

Second, we propose a novel domain-diversity score to quantify the alignment between samples and batch-level distributions using batch normalization statistics and feature maps, enabling real-time evaluation of domain diversity.

Third, we propose a robust TTA framework DATTA for dynamic domain shifts, comprising three key modules: DD, which evaluates domain diversity; DABN, which dynamically aggregates source and test-time statistics for effective normalization; and {DAFT}, which selectively fine-tunes to prevent gradient conflicts, ensuring stable optimization and adaptation.

Finally, extensive experiments reveal that DATTA outperforms baselines across three distinct scenarios, e.g., DATTA achieves the highest average accuracy in the Dynamic-Domain scenario, reaching up to 13\% higher than other state-of-the-art methods at most.

\section{Preliminary}
\label{sec:motivation}

\subsection{Revisiting TTA}

\textbf{Test-time Adaptation.} Let $\mathcal{D}_{\mathcal{S}} = \left\{\mathcal{X}^{\mathcal{S}}, \mathcal{Y}\right\}$ denote the source domain data and $\mathcal{D}_{\mathcal{T}} = \left\{\mathcal{X}^{\mathcal{T}}, \mathcal{Y}\right\}$ denote the target domain data. Each data instance and corresponding label pair $\left(\mathbf{x}_i, y_i\right) \in \mathcal{X}^{\mathcal{S}} \times \mathcal{Y}$ in the source domain follows a distribution $P_{\mathcal{S}}(\mathbf{x}, y)$. Similarly, each target test sample and its label at test time $t$, $\left(\mathbf{x}_t, y_t\right) \in \mathcal{X}^{\mathcal{T}} \times \mathcal{Y}$, follow a distribution $P_{\mathcal{T}}(\mathbf{x}, y)$, with $y_t$ unknown to the learner. The standard covariate shift assumption in domain adaptation is $P_{\mathcal{S}}(\mathbf{x}) \neq P_{\mathcal{T}}(\mathbf{x})$ and $P_{\mathcal{S}}(y \mid \mathbf{x}) = P_{\mathcal{T}}(y \mid \mathbf{x})$. Unlike traditional domain adaptation, which uses pre-collected $\mathcal{D}_{\mathcal{S}}$ and $\mathcal{X}^{\mathcal{T}}$, TTA continuously adapts a pre-trained model $f_\theta(\cdot)$ from $\mathcal{D}_{\mathcal{S}}$ using only the test sample obtained at time $t$.

\textbf{TTA on dynamic stream.} Previous TTA methods typically assume that, at each time $t$, each target sample $\left(\mathbf{x}_t, y_t\right) \in \mathcal{X}^{\mathcal{T}} \times \mathcal{Y}$ is drawn from a time-invariant distribution $P_{\mathcal{T}}(\mathbf{x}, y)$, referred to as single-domain  patterns. However, in many real-world data streams, test-time data streams are inherently dynamic and consist of a mixture of single-domain and multiple-domain  patterns. Specifically, the data may originate from multiple distributions $\{P_{\mathcal{T}}^i\}_{i=1}^M$, where $M$ is the number of domains, representing multiple-domain  patterns. Dynamic-domain data streams interplay between single-domain and multiple-domain patterns which better reflects the complexity of real-world environments. Its data comes from one or multiple distributions. These scenarios continuously evolve and vary across time.

% \subsection{Adaptation in Real-World: Motivations}
% Adapting machine learning models to real-world data stream presents unique challenges due to the dynamic and diverse nature of these environments. Unlike controlled experimental settings presenting by \cite{niu2023towards}, real-world scenarios often involve not only shifts in data distributions but also dynamic diversity patterns, introducing instability in optimization and feature representation. These challenges can be distilled into two core problems, which exhibit markedly different behaviors under high and low-diversity  patterns.

% \begin{figure}
%     \centering
%     \includegraphics[width=1\linewidth]{Definitions/img/tsne.pdf}
%     \caption{Multiple-Domain vs. Single-Domain, t-SNE visualization of ResNet50 features on CIFAR10-C (Severity Level 5), normalized with Batch Normalization (Test-Time Statistics). Different colors represent different classes. }
%     \label{fig:TSNE}
%     % \vspace{-0.38in}
% \end{figure}

\subsection{Motivation}

\textbf{Batch Normalization fails in multiple-domain patterns.} BN improves training stability by normalizing feature distributions using batch-level statistics, but its effectiveness is highly sensitive to the distributions of data within a batch. To gain a deeper understanding of the ability of BN to capture target distributions under multiple-domain patterns, We evaluated the performance of several common BN adaptation methods, including Test-Time BN (TBN), Instance Normalization (IN), IABN~\cite{gong2022note}, and Source Model BN (SBN), as the number of domains gradually increased. As shown in Fig. \ref{fig:motivation-all}(a), every BN method's accuracy significantly drops. This performance deterioration can be attributed to the limited number of samples per domain, which impedes the accurate computation of target distribution statistics (mean and variance). As a result, this problem will make the features' normalization incorrect and lead to distorted feature representations which confuse feature extractions \cite{9879104,jiang2024discoverneighborsadvancedstable,DBLP:conf/nips/ZhaoWC21}. In contrast, single-domain patterns provide a more stable learning environment. In Fig. \ref{fig:motivation-all}(a), each BN method performs well under single-domain patterns. It demonstrates that all samples within a batch originate from a consistent distribution, and BN can compute accurate statistics of target distributions, enabling reliable normalization.

\textbf{Gradients conflict under multiple-domain patterns.} 
% One critical challenge in real-world adaptation is conflicting gradients, which are significantly exacerbated under multiple-domain patterns. 
% When a batch contains samples from multiple heterogeneous distributions (multiple-domain), the gradients computed during optimization often point in opposing directions, creating a \textit{tug-of-war effect}. These conflicts disrupt optimization, destabilize parameter updates, hinder effective generalization across domains and also decrease efficiency.
% When a batch contains samples from multiple heterogeneous distributions (multiple-domain), gradient conflicts can arise, leading to a "tug-of-war effect" that disrupts optimization, causes unstable parameter updates, weakens cross-domain generalization, and reduces efficiency.
To analyze the impact of multiple-domain patterns on model optimization, we measured the L2 norm of gradients produced by the TENT method under both single-domain and multiple-domain settings. As shown in Fig.~\ref{fig:motivation-all}(b), in the multiple-domain setting, gradients from each batch often conflict, leading to a ``tug-of-war effect,'' which results in significant fluctuations in gradient norms. These fluctuations indicate severe conflicts that destabilize the learning process, weaken generalization and reduce efficiency. In contrast, in the single-domain setting, batches consist of samples from similar distributions, resulting in well-aligned gradients. This alignment minimizes conflicts, enabling stable updates and more effective feature learning. Fig.~\ref{fig:motivation-all}(c) compares the accuracy of TENT and NOTE methods before and after parameter updates under multiple-domain patterns, highlighting that fine-tuning in such scenarios can harm adaptation performance.

The previous discussions highlight how multiple domains within the target batch can hinder model adaptation. This naturally raises the question: \textit{How can we identify the diversity among domains and mitigate the impact of such diversity on batch normalization and gradients to achieve better adaptation?}

\section{Proposed Methods}
\begin{figure}[t]
    \centering
    % \begin{adjustwidth}{-\extralength}{0cm}
    \includegraphics[width=0.9\linewidth]{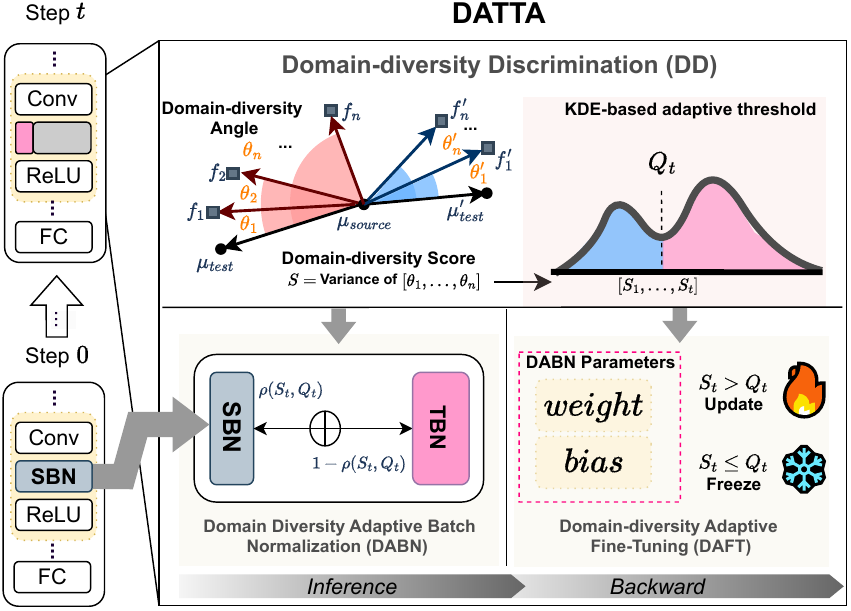}
    % \end{adjustwidth}
    \caption{Overview. DATTA consists of three modules: Diversity Discriminator takes advantage of an Instance-Normalization-guided projection to capture the data features. Based on the discrimination results, DABN and DAFT conduct an adaptive BN re-correcting and model fine-tuning strategy.}
    \label{fig:overview}
    \vspace{-0.25in}
\end{figure}

In this section, building upon our analysis, we address such TTA problem by introducing and proposing a novel DATTA framework. It employs a diversity discrimination module, which effectively utilizes domain diversity metrics to quantify the degree of diversity, then dynamically adjusts BN statistics using enriched source BN and designs adaptive fine-tuning mechanisms. The overall pipeline of DATTA is detailed in Fig.~\ref{fig:overview}.

\subsection{Domain-Diversity Discrimination (DD)}
\label{sec:DS}
% \textbf{Domain-diversity Score.}
% To identify the degree of domain diversity in a domain shift data stream, a natural intuition lies in measuring the alignment between individual samples and the overall batch distribution. Based on this idea, we use the difference between the BN statistics mean of the source domain and the test sample features to represent local shifts. Similarly, the difference between the BN statistics mean of the source domain and the test BN mean represents a global shift. As shown in Fig.~\ref{fig:motivation-all}(d), in the single-domain scenario, the fluctuation between local shift and global shift is very small. In contrast, in the multiple-domain scenario, this fluctuation becomes significantly severe.

To identify the degree of domain diversity in a dynamic domain shift data stream, we propose a domain-diversity score that quantifies the alignment between individual samples and the overall batch distribution. This score combines \textit{local shifts} (differences between individual samples and the training domain) and \textit{global shifts} (differences between the test batch and the training domain). As shown in Fig.~\ref{fig:motivation-all}(d), in single-domain patterns, the fluctuation between local and global shifts is small, whereas in multiple-domain patterns, the fluctuation becomes more severe. This significant difference allows for effective detection and differentiation of multiple-domain and single-domain patterns.

\textbf{Domain-Diversity Score.} By analyzing the deviation angle between local and global shifts, we design a domain diversity-based score to precisely distinguish between multiple-domain and single-domain scenarios. Specifically, the feature map \( f \) is generated by the model's first convolutional layer.  
Let \( \mu_{\text{test}} \) represent the first-layer mean values of the test-time batch normalization statistics, and \( \mu_{\text{source}} \) denote the first-layer mean values of the source model's training batch normalization. Moreover, we introduce the following definition for each sample in a batch:
\label{sec:Domain-diversity andgle}
\newtheorem{definition}{Definition}
\begin{definition}
\textbf{Domain-Diversity Angle}:
The domain-diversity angle \( \theta \) quantifies the difference between the feature vector \( v_f \) and the test distribution vector \( v_t \). It is defined as:
    \[
    \theta = \cos^{-1} \left( \frac{v_f \cdot v_t}{\|v_f\| \|v_t\|} \right),
    \]
here, the local shift vector \( v_f \) represents the difference between the source domain mean and the feature map \( f \): $v_f = \mu_{\text{source}} - f$. 
Similarly, the global shift vector \( v_t \) is defined as the difference between the source domain mean and the test-time batch mean: $v_t = \mu_{\text{source}} - \mu_{\text{test}}$.
\end{definition}

The domain-diversity score \( \mathit{S} \) is then calculated as the variance of all \( \theta \) angles in a batch:
\begin{equation}
    \mathit{S} = \frac{1}{N} \sum_{i=1}^{N} (\theta_i - \overline{\theta })^2,
\end{equation}
where \( \overline{\theta} \) is the mean of all calculated angles \( \theta_i \) within the batch, and \( N \) denotes the total number of samples in the batch. A larger \( \mathit{S} \) indicates higher domain diversity (i.e., multiple-domain samples), while a smaller \( \mathit{S} \) reflects lower diversity (i.e., single-domain samples). By setting a threshold, we can distinguish between single-domain and multiple-domain batches.

% The larger the value of the score, the more diverse the domains in this batch. Conversely, the smaller the value, the fewer the domains contained in this batch. By setting a reasonable threshold, it will accurately distinguish between single-domain and multiple-domain.

% This method allows us to effectively measure the diversity of data distribution conditions in batch processing. As shown in {Fig. \ref{fig:motivation-all}(c)}, this method enables to quantify of the extent of domain variability, the domain diversity score provides a robust metric for analyzing both single-domain and multiple-domain conditions. 

\textbf{KDE-based adaptive threshold.}
\label{sec:DD} 
The distribution of the score value also changes dynamically according to the changes in the data stream. Therefore, this threshold must also adapt dynamically. Drawing inspiration from the ability of Kernel Density Estimation (KDE)~\cite{wkeglarczyk2018kernel} to uncover the underlying structure of data by smoothing distributions—where the valleys naturally represent boundaries between distinct clusters-we developed a method leveraging probability density estimation to precisely identify separation threshold \( Q_t \) between different single-domain batches and multiple-domain batches.
% \todo{add ref}

At test step \( t \), we have a KDE-based domain diversity score probability density function \( k(\mathit{S_t}) \) from the historical domain-diversity scores:
\begin{equation}
  k(\mathit{S_t}) = \frac{1}{n h \sqrt{2 \pi}} \sum_{i=1}^{n} \exp\left(-\frac{\left(\mathit{S_t} - \mathit{S}_i\right)^2}{2 h^2}\right), \quad \mathit{S}_i \in \mathcal{S},
\end{equation}
where \( h = 1.06 \sigma n^{-\frac{1}{5}} \) is the bandwidth parameter determined by Scott's rule of thumb \cite{scott2010scott}, \( n \) is the size of the historical scores set \( \mathcal{S} \), and \( \sigma \) is their standard deviation.

% At each test step \( t \), we first compute the domain diversity score \( \mathit{S_t} \) for the current batch. Then, using the historical domain diversity scores set \(\mathcal{S} = \{\mathit{S}_1, \mathit{S}_2, \ldots, \mathit{S}_i, \ldots, \mathit{S}_{t-1}\}\), collected from the previous \( t-1 \) steps. Then we have a KDE-based domain diversity score probability density function \( k(\mathit{S_t}) \), defined as:
% \begin{equation}
%   k(\mathit{S_t}) = \frac{1}{n h \sqrt{2 \pi}} \sum_{i=1}^{n} \exp\left(-\frac{\left(\mathit{S_t} - \mathit{S}_i\right)^2}{2 h^2}\right), \quad \mathit{S}_i \in \mathcal{S},  
% \end{equation}
% where \( h = 1.06 \sigma n^{-\frac{1}{5}} \) is the bandwidth parameter, determined using Silverman's rule of thumb,  a widely used method for selecting the bandwidth in KDE. Here, \( n \) denotes the size of the domain diversity scores set \(\mathcal{S}\), and \( \sigma \) represents their standard deviation.

The peaks in the KDE represent regions where the scores are concentrated. The height of a peak reflects the density of the score data in that region. Different peaks may correspond to different score patterns. Valleys, on the other hand, represent regions where the scores are sparse. The depth of a valley reflects the sparsity of the score data in that region. A valley may indicate a separation or discontinuity in the score distribution. 
% To compute \( Q_{\text{valley}} \), we evaluate \( k(S_i) \) for all \( S_i \) in the interval \([S_{\text{$pk$}_1}, S_{\text{$pk$}_2]}\) and find the minimum value:
To compute the valley \( Q_{\text{valley}} \), we find the domain-diversity score with minimum density between these peaks:
\begin{equation}
    Q_{\text{valley}}  = \argmin_{S_i \in [S_{\text{pk}_1}, S_{\text{pk}_2}]} k(S_i), \mathit{S}_i \in \mathcal{S},  
\end{equation}
where \( S_{\text{$pk$}_1} \) and \( S_{\text{$pk$}_2} \) are the scores of the two largest density peaks, \(\text{$pk$}_1\) and \(\text{$pk$}_2\), respectively. \( k(S_i) \) denotes the KDE function, which estimates the probability density at a given point \( S_i \). This \( Q_{\text{valley}} \) separates multiple-domain batches (\(\mathit{S}_t > Q_{\text{valley}}\)) from single-domain batches (\(\mathit{S}_t \leq Q_{\text{valley}}\)). 

To improve robustness, we further adjust the threshold \( Q_t \) based on the relative dominance of the two peaks. Let \( k(S_{\text{pk}_1}) \geq k(S_{\text{pk}_2}) \), and define their ratio as:
\begin{equation}
     r_{\text{pk}} = \frac{k(S_{\text{pk}_1})}{k(S_{\text{pk}_2})}, 
\end{equation}
% $r_{\text{pk}}$ quantifies the relative dominance of the two largest peaks in the score distribution. A lower $r_{\text{pk}}$ indicates a more balanced distribution, suggesting the presence of multiple-domain samples, where the threshold $Q_t$ is set to $Q_{\text{valley}}$ to separate single- and multiple-domain batches. $n_{\text{pk}}$ represents the threshold for assessing balance in domain diversity score distributions. When the ratio $r_{\text{pk}}$ falls below $n_{\text{pk}}$, it indicates a higher balance.
$r_{\text{pk}}$ quantifies the relative dominance of the two largest peaks in the score distribution. $n_{\text{pk}}$ represents the threshold for assessing balance in domain-diversity score distributions. When $r_{\text{pk}}$ falls below $n_{\text{pk}}$, it indicates a more balanced distribution, suggesting the presence of single-domain samples, where the threshold $Q_t$ is set to $Q_{\text{valley}}$ to separate single- and multiple-domain batches. If the peaks are balanced (\( r_{\text{pk}} \leq n_{\text{pk}} \)), the threshold remains \( Q_t = Q_{\text{valley}} \). Otherwise, it is set to zero to handle highly imbalanced distributions:
\begin{equation}
     Q_t = Q_{\text{valley}} \cdot \mathbb{I}(r_{\text{pk}} \leq n_{\text{pk}}).
\end{equation}

This adaptive threshold mechanism ensures accurate discrimination between single-domain and multiple-domain scenarios, even under dynamically shifting data distributions. 
% If the ratio \( r_{\text{pk}} \) exceeds a predefined threshold \( n_{\text{pk}} \), indicating a highly imbalanced distribution, the \( Q_t \) is adjusted to an infinitesimally small value (\( Q_t = 0 \)) to prioritize the inclusion of all single-domain samples. Otherwise, the threshold is determined by the valley \( Q_{\text{valley}} \) between the two peaks. This adjustment is expressed as:
% \begin{equation}
%     \begin{aligned}
%     Q_t &= 
%         \begin{cases} 
%         Q_{\text{valley}}, & \text{if } r_{\text{pk}} \leq n_{\text{pk}}, \\
%         0, & \text{if } r_{\text{pk}} > n_{\text{pk}}.
%         \end{cases},
%     \enspace\text{where}\enspace r_{\text{pk}} = \frac{k(S_{\text{pk}_1})}{k(S_{\text{pk}_2})}.
%     \end{aligned}
% \end{equation}

% This mechanism ensures that the method remains robust under extreme data distributions, where one mode dominates the domain diversity score distribution. By dynamically adjusting the threshold \( Q_t \), the model can effectively balance the trade-off between the source and test-time statistics, improving its adaptability to real-world data stream with varying domain diversity.

\begin{table*}[t]
    \vspace{-0.1in}
    \caption{
    Comparison of state-of-the-art methods on CIFAR-10-C (C10-C), CIFAR-100-C (C100-C) and ImageNet-C (IN-C) using EfficientViT-M5 at severity level 5 with a \textsc{Batch Size} of 64, under Single-Domain, Multiple-Domain and Dynamic-Domain scenarios, evaluated by Accuracy (\%). Bold indicates the best result, and \underline{underlining} denotes the second-best.  
    % Comparison of state-of-the-art methods on CIFAR10-C and CIFAR100-C using EfficientViT-M0, and ImageNet-C using EfficientViT-M5 at severity level 5 with a \textbf{\textsc{Batch Size of 64}}, under \textbf{High Diversity} and \textbf{Real Diversity} scenarios, evaluated by \textbf{Accuracy (\%)}. \textbf{Bold} indicates the best result, and \underline{underlining} denotes the second-best.
    % Comparison of state-of-the-art methods on CIFAR10-C, CIFAR100-C, and ImageNet-C at severity level 5 with a \textbf{\textsc{Batch Size of 64}} under Dynamic and Dynamic-S scenarios on EfficientViT, evaluated by \textbf{Accuracy (\%)}. The  \textbf{bold} value signifies the best result, and the second best accuracy is \underline{underlined}.
    }
    \vspace{-0.15in}
    % \label{tab:imagenet-c-normal}
\newcommand{\tabincell}[2]{\begin{tabular}{@{}#1@{}}#2\end{tabular}}
 \begin{center}
  \begin{threeparttable}
 % \normalsize
    \resizebox{1.0\linewidth}{!}{
 	\begin{tabular}{l|l|cccc|cccc|cccc|c}
\toprule
 	\multicolumn{2}{c}{}  & \multicolumn{4}{c}{Single-Domain}  & \multicolumn{4}{c}{Multiple-Domain}  & \multicolumn{4}{c}{Dynamic-Domain}  &\multicolumn{1}{c}{}\\
  \cmidrule{1-15}
 	  Method &Venue & {C10-C} &{C100-C}& {IN-C}& ~Avg.~$\uparrow$~ & {C10-C} &{C100-C}& {IN-C}& ~Avg.~$\uparrow$~  & {C10-C} &{C100-C}& {IN-C} & ~Avg.~$\uparrow$~  & ~Avg-All$\uparrow$~ \\
     \cmidrule{1-15}
    Source & CVPR'23 & 74.63 & 42.88 & \underline{27.56} & 48.36 & 74.63 & 42.88 & \underline{27.65} & 48.39 & 74.63 & 42.87 & \underline{27.42} & 48.31 & 48.35 \\
    % BN Stats  &ICLR'21 & & & & & 75.42 & 44.55 & 28.07 & & 78.60 & \underline{41.62} &  & &\\
    TENT  &CVPR'21 & 81.66 & 48.49 & 25.47 & 51.87 & 75.29 & 42.23 & 20.33 & \underline{45.95} & 78.77 & \underline{45.20} & 23.09 & \underline{49.02} & 48.95 \\
    % EATA  & ICML'22 & 81.22 & 48.32 & 25.93 & 51.82 & 74.82 & 42.77 & 20.45 & 46.01 & 78.02 & 45.58 & 23.06 & 48.89 & 48.91 \\
    NOTE  &NIPS 22 & 76.53 & 35.68 & 8.98 &40.40 & 74.11 & 34.43 & 8.85 & 39.13& 75.08 & 34.58 & 8.12 &39.26 &39.59\\
    SAR  &ICLR'23 & 81.67 & 47.82 & 26.19 & 51.89 & 75.26 & 41.84 & 20.47 & 45.86 & 78.74 & 44.23 & 23.10 & 48.69 & 48.81 \\
    RoTTA  &CVPR'23 & 81.72 & \underline{49.82} & 27.18 & \underline{52.91} & \textbf{76.20} & \underline{43.39} & 20.62 & 46.74 & 77.80 & 44.83 & 23.12 & 48.58 & \underline{49.41} \\
    ViDA  &ICLR'24 & 81.22 & 48.07 & 25.55 & 51.61 & 74.82 & 41.65 & 7.50 & 41.33 & 78.01 & 44.86 & 14.03 & 45.64 & 46.19 \\
    DeYO &ICLR'24 & \underline{81.92} & 45.29 & 26.05 & 51.09 & \underline{76.13} & 40.16 & 20.20 & 45.50 & \textbf{79.60} & 40.30 & 23.41 & 47.77 & 48.12 \\
    \rowcolor{cyan!15}Ours &Proposed& \textbf{82.16} & \textbf{50.77} & \textbf{37.07} & \textbf{56.66} & 75.67 & \textbf{44.80} & \textbf{28.32} & \textbf{49.59} & \underline{78.94} & \textbf{47.67} & \textbf{30.74} & \textbf{52.45} & \textbf{52.90} \\
    \bottomrule
                \end{tabular} 
      }
    \end{threeparttable}
    \end{center}
\label{tab:main}
\vspace{-0.25in}
\end{table*}

\subsection{Domain-Diversity Adaptive Batch Normalization (DABN)}
\label{sec:DABN}
To address the challenges posed by dynamic domain shift TTA, we propose DABN, a novel normalization method that dynamically adjusts its strategy based on domain-diversity scores. This adaptive strategy allows DABN to seamlessly transition between single-domain and multiple-domain scenarios. 

As described in \S Sec.~\ref{sec:motivation}, recent TTA algorithms rely solely on re-calculating BN statistics. However, these BN-based methods suffer from significant performance degradation when handling data streams with increasing domain diversity. This is because standard BN assumes consistent data distributions within a batch, which becomes invalid in multiple-domain or highly diverse scenarios. 

To mitigate this issue, DABN adaptively balances the use of source domain statistics (\(\mu_{\text{source}}, \sigma^2_{\text{source}}\)) and current batch statistics (\(\mu_{\text{test}}, \sigma^2_{\text{test}}\)) based on a domain-diversity score (\(\mathit{S}_t\)). Specifically, when high domain diversity is detected (e.g., multiple-domain batches), DABN increases reliance on the source domain statistics to reduce errors caused by inaccurate batch estimates. Conversely, in low diversity scenarios, DABN primarily utilizes current batch statistics for better adaptation to the target distribution. 
The computation of DABN is governed by the following equations:
\begin{equation}
\rho = \alpha_{\text{single}} \cdot \mathbb{I}(S_t < Q_t) + \alpha_{\text{multi}} \cdot \mathbb{I}(S_t \geq Q_t),
\end{equation}
\begin{equation}
\mu_{\text{DABN}} = \rho \cdot \mu_{\text{source}} + (1 - \rho) \cdot \mu_{\text{test}},
\end{equation}
\begin{equation}
\sigma_{\text{DABN}}^{2} = \rho \cdot \sigma_{\text{source}}^{2} + (1 - \rho) \cdot \sigma_{\text{test}}^{2},
\end{equation}
where \(\alpha_{\text{single}}\) and \(\alpha_{\text{multi}}\) are predefined weight coefficients used for single-domain and multiple-domain scenarios, respectively, and \(\mathit{S}_t\) denotes the domain-diversity score. The threshold \(Q_t\) determines whether a batch is treated as single-domain or multiple-domain. 
% When \(\mathit{S}_t\) exceeds \(Q_t\), DABN prioritizes source domain statistics for more stable normalization. Otherwise, it relies more heavily on current batch statistics for better adaptability. 
Finally, the complete formulation of DABN is expressed as:
\begin{equation}
\textbf{DABN}:= \gamma \cdot \frac{\mathbf{f} - \mu_{\text{DABN}}}{\sqrt{\sigma_{\text{DABN}}^{2} + \epsilon}} + \beta,
\end{equation}
where \(\mathbf{f}\) is the input feature, \(\gamma\) and \(\beta\) are the learnable scaling and shifting parameters, and \(\epsilon\) is a small constant for numerical stability.

\subsection{Domain-Diversity Adaptive Fine-Tuning (DAFT)}
\label{sec:DaFT}
After updating the BN layer's statistical values, the model's affine parameters need to be fine-tuned to adapt to the target domain.  However, our findings indicate that when the batch data exhibits high domain diversity, gradient updates may suffer from instability due to conflicting optimization directions. As a result, these updates can become ineffective, or even detrimental to model performance. 

To address this issue, we propose DAFT, which selectively applies parameter updates only when the batch data has a low domain-diversity score. This ensures that the model avoids wasteful or harmful adjustments caused by highly diverse data distributions. Specifically, the loss function is defined as follows:
\begin{equation}
\mathcal{L} = \mathbb{I}_{\left\{S_t \geq Q_t\right\}} \operatorname{Ent}_{\boldsymbol{\theta}}(\mathbf{x}),
\end{equation}
where \(\operatorname{Ent}_{\boldsymbol{\theta}}(\mathbf{x})\) is the cross-entropy loss, $\mathbf{x}$ is the model input, and \(\mathbb{I}_{\left\{S \geq Q\right\}}\) is the indicator function that equals 1 if the domain-diversity score \(S\) is greater than the threshold \(Q\), and 0 otherwise.

\section{Experiments}
\subsection{Experimental Setup}
We implemented the DATTA method as well as baseline methods within the TTAB framework \cite{zhao2023ttab}, a widely used TTA benchmark. Below, we provide detailed information on the datasets, models, and experimental configurations. 

\textbf{Datasets and Models.}
To evaluate the robustness of our method against corrupted data, we used three standard benchmarks: CIFAR-10-C, CIFAR-100-C, and ImageNet-C \cite{hendrycks2019robustness}. Each dataset includes 15 corruption types (e.g., Gaussian noise, defocus blur) across five severity levels, simulating real-world visual challenges. For rigorous evaluation, we focused on the most severe corruption level (level 5) from each type.

We tested our method on two representative architectures: EfficientViT \cite{liu2023efficientvit}, a Vision Transformer designed for efficient computation, and ResNet-50 \cite{heDeepResidualLearning2015}, a widely-used convolutional network with Batch Normalization, aligning closely with our adaptive normalization approach.

\textbf{Adaptation Scenarios.} In our experiments, we utilized three scenarios: Single-Domain, Multiple-Domain and Dynamic-Domain. In the Single-Domain scenario, each batch of input samples is exclusively from one domain. In the Multiple-Domain scenario, each batch of input samples is composed of data from a randomly selected set of several different domains. In the Dynamic-Domain scenario, each batch consists of i.i.d. samples, sourced from either multiple domain or a single domain. 

 % dynamic-domain shift data streams，single-domain batches （pattern）, multi-domain batches （pattern）

\textbf{{Baselines.}} In our experiments, we present the results in terms of top-1 accuracy, comparing our method with various cutting-edge TTA methods. These include TENT \cite{wangTentFullyTesttime2021b}, NOTE \cite{gong2022note}, SAR \cite{niu2023towards}, RoTTA \cite{yuanRobustTestTimeAdaptation2023}, ViDA \cite{liu2023vida}, and DeYO \cite{lee2024entropy}.

\textbf{{Hyperparameter Configurations.} }
The hyperparameters are divided into two categories: those shared by all baselines and those specific to the method.  
(1) The shared hyperparameters are as follows: The optimizer is set to SGD. The learning rate (LR) for CIFAR-10-C and CIFAR-100-C is set to $0.0001$, while the LR for ImageNet-C is reduced to $0.00001$. The batch size is fixed at $64$.  
(2) For DABN, after comprehensive evaluation of various parameter combinations, we set $\alpha_{\text{single}}$ to 0.6 for the Single-Domain scenario and $\alpha_{\text{multi}}$ to 0.85 for the Multiple-Domain scenario, as these values yielded optimal performance.
% (2) Specifically, for DABN, $\alpha_{\text{single}}$ is set to 0.6 in the Single-Domain scenario, and $\alpha_{\text{multi}}$ is set to 0.85 in the Multiple-Domain scenario.
%$\alpha$ is a hyperparameter that determines the adjustment level based on the current batch information, and it is set to $0.2$.

%After the test data are input into the model, all data is first forwarded once to obtain the inference result.
% Then, the model adapts according to the pseudo-labels obtained from the inference, with the number of epochs set to 1, i.e., one backward propagation.
% 2) The hyperparameters specific to each method are set according to the following references: the hyperparameters for TBN follow the settings in \cite{wangTentFullyTesttime2021b}; the hyperparameters for IABN are based on the settings in \cite{gongNOTERobustContinual2023}; and the hyperparameters for $\alpha$-BN also follow the settings in \cite{wangTentFullyTesttime2021b}.

\begin{table}[t]
    % \vspace{-0.1in}
    \caption{
    Comparison of memory (GB) and latency (s) for ImageNet-C on a V100 GPU with EfficientViT-M5. Bold indicates best results excluding Source model.% \textbf{The first row represents Memory, and the second row represents Latency.}% CIFAR-10-C, CIFAR-100-C, 
    % \textbf{\textsc{Batch Size of 64}} under Dynamic-S scenarios, evaluated by \textbf{Latency (s)}.
    }
    \vspace{-0.15in}
\newcommand{\tabincell}[2]{\begin{tabular}{@{}#1@{}}#2\end{tabular}}
 \begin{center}
  \begin{threeparttable}
 % \LARGE
% \scriptsize
    \resizebox{\linewidth}{!}{
  \begin{tabular}{c|c|ccccccc}
 	% \multicolumn{2}{c}{}  & \multicolumn{3}{c}{Latency}\\
  % \cmidrule{3-5}
  \toprule
   &Source & TENT & NOTE & SAR & RoTTA & ViDA & DeYO &  Ours\\
     \midrule
Mem. & 0.34 & 1.39 & 2.71 & 1.39 & 3.56 & 2.66 & 1.62 & \cellcolor{cyan!15}\textbf{0.93}\\
Lat.&0.03 & 0.19 & 5.37 & 0.26 & 0.65 & 4.58 & 0.22 & \cellcolor{cyan!15}\textbf{0.19}\\
    \bottomrule
                \end{tabular} 
        }
    \end{threeparttable}
    \end{center}
\label{tab:efficiency}
\vspace{-0.35in}
\end{table}

% \begin{table}[t]
%     % \vspace{-0.1in}
%     \caption{
%     Comparison of memory (GB) and latency (s) for ImageNet-C on a V100 GPU with EfficientViT-M5. Bold indicates best results excluding Source model.% CIFAR-10-C, CIFAR-100-C, 
%     % \textbf{\textsc{Batch Size of 64}} under Dynamic-S scenarios, evaluated by \textbf{Latency (s)}.
%     }
%     \vspace{-0.1in}
% \newcommand{\tabincell}[2]{\begin{tabular}{@{}#1@{}}#2\end{tabular}}
%  \begin{center}
%   \begin{threeparttable}
%  % \LARGE
% % \scriptsize
%     % \resizebox{0.6\linewidth}{!}{
%   \begin{tabular}{l|l|cc}
%  	% \multicolumn{2}{c}{}  & \multicolumn{3}{c}{Latency}\\
%   % \cmidrule{3-5}
%  	  % Method &\hspace{5pt}Venue\hspace{5pt}  &{Memory (GB)}& {Latency (s)}  \\
%       \toprule
%       Method &Venue &{Memory}& {Latency}  \\
%      \midrule
%     Source  &CVPR'23 &0.34 &0.03\\
%     % BN Stats  &\hspace{5pt}ICLR'21\hspace{5pt} &0.018 & 0.018 & 0.068  \\
%          \cmidrule {1-4}
%     TENT  &CVPR'21   & 1.39 & 0.19  \\
%     % EATA  &\hspace{5pt}ICML'22\hspace{5pt}  & 1.44 & 0.20 \\
%     NOTE  &NIPS'22   & 2.71 & 5.37  \\
%     % CoTTA  &\hspace{5pt}CVPR'22 & 0.543 & 0.541 & 5.322  \\
%     SAR  &ICLR'23   & 1.39 & 0.26  \\
%     RoTTA  &CVPR'23& 3.56 & 0.65  \\
%     ViDA  &ICLR'24  & 2.66 & 4.58 \\
%     DeYO  &ICLR'24& 1.62 & 0.22 \\
%     \rowcolor{cyan!15}Ours &Proposed & \textbf{0.93} & \textbf{0.19}\\  
%     \bottomrule
%                 \end{tabular} 
%         % }
%     \end{threeparttable}
%     \end{center}
% \label{tab:efficiency}
% \vspace{-0.2in}
% \end{table}

\begin{table}[t]
    % \vspace{-0.1in}
    \caption{
    Comparison of methods' accuracy (\%) on CIFAR-100-C using ResNet-50 at severity level 5 with a \textsc{Batch Size of 64}, under Single-Domain, Multiple-Domain and Dynamic-Domain scenarios. Bold indicates the best result.
    }
    \vspace{-0.15in}
\newcommand{\tabincell}[2]{\begin{tabular}{@{}#1@{}}#2\end{tabular}}
 \begin{center}
  \begin{threeparttable}
 % \LARGE
% \scriptsize
    % \resizebox{0.6\linewidth}{!}{
  \begin{tabular}{l|l|cccc}
 	% \multicolumn{2}{c}{}  & \multicolumn{3}{c}{Latency}\\
  % \cmidrule{3-5}
  \toprule
 	  Method &Venue&{Single} &{Multiple}& {Dynamic} &~Avg.~$\uparrow$~  \\
     \midrule
    Source  &CVPR'16 & 28.59 &28.59 &28.59 &28.59\\
    % BN Stats  &\hspace{5pt}ICLR'21\hspace{5pt} &0.018 & 0.018 & 0.068  \\
         % \cmidrule {1-4}
    TENT  &CVPR'21  & 50.48 & 32.13 & 39.47 &40.69 \\
    % EATA  &\hspace{5pt}ICML'22\hspace{5pt}  & 1.44 & 0.20 \\
    NOTE  &NIPS'22  &26.38 & 24.55 & 26.38&25.77  \\
    % CoTTA  &\hspace{5pt}CVPR'22\hspace{5pt}  & 0.543 & 0.541 & 5.322  \\
    SAR  &ICLR'23  & 49.67 & 31.44 & 40.18 &40.43 \\
    RoTTA  &CVPR'23&39.20 &22.18 & 28.85 &30.08 \\
    ViDA  &ICLR'24 & 46.80 & 32.14& 39.47 & 39.47 \\
    DeYO  &ICLR'24& \textbf{51.45} & 30.35 & 33.87 &38.56\\
    \rowcolor{cyan!15}Ours &Proposed &50.03& \textbf{33.52} & \textbf{41.11}& \textbf{41.55}\\  
    \bottomrule
                \end{tabular} 
        % }
    \end{threeparttable}
    \end{center}
\label{tab:resnet}
\vspace{-0.25in}
\end{table}

\subsection{Robustness under Dynamic Domain Shift Data Stream}
% We performed experiments in two separate scenarios: High Diversity and Dynamic-Domain. %In alignment with the configurations used in prior studies, we selected the most severely corrupted samples (level 5) from each type of corruption.

\textbf{Overall results.} Tab.~\ref{tab:main} reports the performance of various TTA methods across three adaptation scenarios. Our method consistently outperforms all baselines, achieving accuracies of $56.66\%$, $49.59\%$, and $52.45\%$ in the Single-, Multiple-, and Dynamic-Domain scenarios, respectively. Overall, our approach achieves the best Avg-All accuracy of $52.90\%$, which is $3.49\%$ higher than the second-best method (RoTTA, $49.41\%$) and $13.31\%$ higher than the worst performing method (NOTE, $39.59\%$). These significant improvements underline the robustness and generalization capability of our method across a wide range of domain adaptation challenges.

% Tab.~\ref{tab:main} displays the performance of various TTA methods in Multiple-Domain and Dynamic-Domain scenarios. It is clear from the table that our approach significantly outperforms other benchmarks in terms of average accuracy across the two scenarios. Our method shows a considerable advantage, achieving an average accuracy approximately 13\% higher than the lowest-performing method (NOTE) and about 3\% higher than the second-performing method (Source), indicating its robustness in both single and multiple domain contexts. 

% Tab.~\ref{tab:main} displays the performance of various TTA methods in Multiple-Domain and Dynamic-Domain scenarios. It is clear from the table that our approach significantly outperforms other benchmarks in terms of average accuracy across the two scenarios. In the Multiple-Domain scenario, our method shows a considerable advantage, achieving an average accuracy approximately 11\% higher than the lowest-performing method (NOTE) and about 2\% higher than the second-performing method (Source), demonstrating its strength in handling batch data across multiple domains. In the Dynamic-Domain scenario, our average accuracy is around 11\% higher than the lowest-performing method (NOTE) and approximately 3\% higher than the second-performing method (Source), indicating its robustness in both single and multiple domain contexts. 

% This underscores the effectiveness of our method in handling the static and dynamic patterns.

\textbf{Efficiency.} Tab.~\ref{tab:efficiency} compares the memory usage and latency of methods. Our method shows competitive efficiency. It utilizes only 0.93 GB of memory, which is significantly lower than the memory consumption of RoTTA (3.56 GB) and ViDA (2.66 GB). And our method achieves a latency of 0.19 seconds, substantially lower than NOTE (5.37 seconds) and ViDA (4.58 seconds). Although Source (0.34 GB and 0.03 seconds) has the best efficiency, our method still outperforms most benchmarks. These results demonstrate that our method provides a strong balance between efficiency and high performance.

%, making it suitable for real-time applications where both efficiency and accuracy are crucial.
% This balance enhances QoE, ensuring optimal service performance and satisfaction for end-users.Although the Source model has the lowest memory usage at 0.34 GB, our method still outperforms most of the other benchmarks. 

\textbf{Robust across different model architectures.} Tab.~\ref{tab:resnet} presents the comparison of TTA methods in the ResNet-50 under three adaptation scenarios. Our method achieves the highest average accuracy of 41.55\%, which is approximately 16\% higher than the worst performing method (NOTE) and about 1\% higher than the second best method (TENT). These results indicate cross-model applicability. 
% These results indicate improvements that are applicable across various models.

\begin{figure}[t]
    \centering
    \includegraphics[width=1\linewidth]{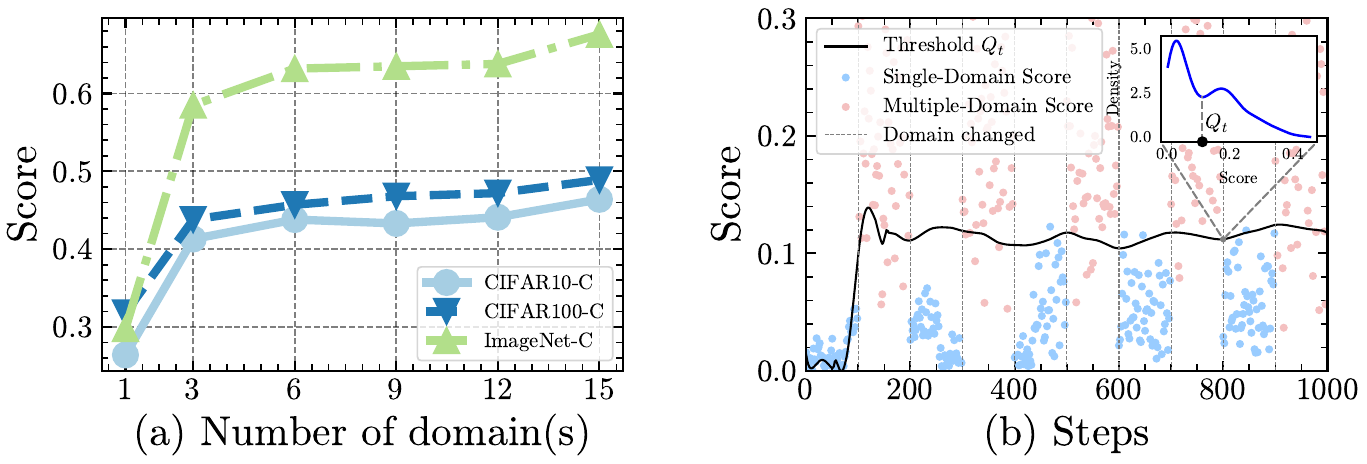}
    \vspace{-0.15in}
    \caption{\textbf{(a)} Illustration of Domain-Diversity Score with growing number of domains by using ResNet-50. \textbf{(b)}  Steps vs. Domain-Diversity Score by using ImageNet-C at severity level 5 and ResNet-50. The upper right inset is Domain-Diversity Score in Gaussian KDE Model.}
    \label{fig:enter-label}
    \vspace{-0.25in}
\end{figure}

\textbf{The effectiveness of the domain-diversity score and the threshold.} Fig.~\ref{fig:enter-label}(a) demonstrates DATTA's performance across multiple datasets, where the score increases with the number of domains, indicating that the designed score effectively captures domain diversity. Fig.~\ref{fig:enter-label}(b) illustrates how the threshold dynamically adjusts with domain transitions, showcasing the adaptability of the method to changes in the number of domains.

\textbf{Ablation Study.} 
This study evaluates the contributions of DD, DAFT, and DABN modules to ImageNet-C using the EfficientViT-M5 in three adaptation scenarios, which detailed in Tab.~\ref{tab:ablation}. The DD module cannot be used alone, so combinations with DAFT and DABN were tested. TENT achieves 22.96\% average accuracy as the baseline. Adding the DAFT module (DD+DAFT) improves average accuracy to 25.24\%, while replacing DAFT with DABN (DD+DABN) achieves a higher average accuracy of 30.13\%. The full combination (DD+DAFT+DABN) gives the highest average accuracy of 32.43\%, outperforming all other setups. These results highlight the complementary roles of DAFT and DABN modules, and demonstrate that integrating all modules significantly enhances robustness and performance under diverse conditions.
\begin{table}[tbp]
\centering
\caption{Ablation study of DD, DAFT and DABN modules (DD can not be used alone). \textbf{Bold} indicates best results. } % 你可以在这里修改表格标题
\vspace{-0.1in}
\resizebox{\columnwidth}{!}{
\begin{tabular}{r|ccc|c}
\toprule
    \multicolumn{1}{r}{}&   \multicolumn{1}{c}{Single} & \multicolumn{1}{c}{Multiple}  &  \multicolumn{1}{c}{Dynamic}  & \multicolumn{1}{c}{Avg-Acc. $\uparrow$ } \\
    % \cmidrule{1-6}
    % \multicolumn{1}{c}{Method} &  train & DD & train & DD & \\
    \cmidrule{1-5}
    Fully Test-Time Adaptation (TENT) & 25.47 & 20.33 & 23.09 & 22.96 \\
    DD+DAFT & 23.22 & 28.10 & 24.39 & 25.24 \\
    DD+DABN & 36.69 & 24.70 & 29.01 & 30.13 \\
    \rowcolor{cyan!15}DD+DABN+DAFT (Ours) & \textbf{37.20} & \textbf{29.43} & \textbf{30.67} & \textbf{32.43} \\
    \bottomrule
 \end{tabular}
 }
\label{tab:ablation}%
\vspace{-0.25in}
\end{table}
\section{Conclusion}
% This paper presents DATTA, a framework for handling dynamic domain shifts in test-time adaptation. By leveraging a domain diversity score, DABN, and DAFT, it effectively mitigates normalization errors and gradient conflicts. Experiments show DATTA outperforms existing methods in accuracy and efficiency across various scenarios.
This paper introduces DATTA, a framework for test-time adaptation to dynamic domain shifts data streams. It uses the Domain-Diversity Score for dynamic domains recognition and combines adaptive batch normalization with fine-tuning to reduce normalization errors and gradient conflicts across multiple domains. Experiments show DATTA significantly outperforms existing methods in accuracy and efficiency, offering a novel solution for dynamic domain adaptation during testing.
\bibliographystyle{splncs04}
\bibliography{main}

@String(CVPR= {IEEE Conf. Comput. Vis. Pattern Recog.})

@String(ICLR = {Int. Conf. Learn. Represent.})

@String(CVPR  = {CVPR})

@String(ICLR  = {ICLR})

@InProceedings{liu2023efficientvit,
    title     = {EfficientViT: Memory Efficient Vision Transformer with Cascaded Group Attention},
    author    = {Liu, Xinyu and Peng, Houwen and others},
    booktitle = {CVPR},
    year      = {2023},

}

@inproceedings{
    lee2024entropy,
    title={Entropy is not Enough for Test-time Adaptation: From the Perspective of Disentangled Factors},
    author={Jonghyun Lee and Dahuin Jung and others},
    booktitle={ICLR},
    year={2024},
}

@inproceedings{hendrycks2019robustness,
  title={Benchmarking Neural Network Robustness to Common Corruptions and Perturbations},
  author={Dan Hendrycks and Thomas Dietterich},
  booktitle={ICLR},
  year={2019}
}

@inproceedings{recht2019imagenet,
  title={Do imagenet classifiers generalize to imagenet?},
  author={Recht, Benjamin and Roelofs, Rebecca and others},
  booktitle={ICML},
  year={2019},
}

@inproceedings{choi2021robustnet,
  title={Robustnet: Improving domain generalization in urban-scene segmentation via instance selective whitening},
  author={Choi, Sungha and others},
  booktitle={CVPR},
  year={2021}
}

@inproceedings{niu2023towards,
  title={Towards Stable Test-Time Adaptation in Dynamic Wild World},
  author={Niu, Shuaicheng and Wu, Jiaxiang and others},
  booktitle = {ICLR},
  year = {2023}
}

@inproceedings{yuanRobustTestTimeAdaptation2023,
  title={Robust test-time adaptation in dynamic scenarios},
  author={Yuan, Longhui and Xie, Binhui and others},
  booktitle={CVPR},
  year={2023}
}

@inproceedings{wangContinualTestTimeDomain2022,
  title={Continual test-time domain adaptation},
  author={Wang, Qin and others},
  booktitle={CVPR},
  year={2022}
}

@inproceedings{zhao2023ttab,
  title     = {On Pitfalls of Test-time Adaptation},
  author    = {Zhao, Hao and others},
  booktitle = {ICML},
  year      = {2023},
}

@inproceedings{heDeepResidualLearning2015,
  title={Deep residual learning for image recognition},
  author={He, Kaiming and others},
  booktitle = {CVPR},
  year={2016}
}

@inproceedings{wangTentFullyTesttime2021b,
  title={Tent: Fully test-time adaptation by entropy minimization},
  author={Wang, Dequan and Shelhamer, Evan and others},
  booktitle = {ICLR},
  year={2021}
}

@inproceedings{gong2022note,
    author = {Gong, Taesik and Jeong, Jongheon and others},
    title = {{NOTE}: Robust Continual Test-time Adaptation Against Temporal Correlation},
    booktitle = {NeurIPS},
    year = {2022}
}

@inproceedings{liu2023vida,
  title={Vida: Homeostatic visual domain adapter for continual test time adaptation},
  author={Liu, Jiaming and Yang, Senqiao and others},
  booktitle={ICLR},
  year={2024}
}

@INPROCEEDINGS{9879104,
  author={Huang, Lei and Zhou, Yi and others},
  booktitle={CVPR}, 
  title={Delving into the Estimation Shift of Batch Normalization in a Network}, 
  year={2022},
}

@misc{jiang2024discoverneighborsadvancedstable, 
      title= {Discover Your Neighbors: Advanced Stable Test-Time Adaptation in Dynamic World}, 
      author={Qinting Jiang and Chuyang Ye and others},
      year={2024},
      eprint={2406.05413},
      archivePrefix={arXiv},
      primaryClass={cs.LG},
}

@inproceedings{DBLP:conf/nips/ZhaoWC21,
  author       = {Yin Zhao and
                  Minquan Wang and
                  others},
  title        = {Reducing the Covariate Shift by Mirror Samples in Cross Domain Alignment},
  booktitle    = {NeurIPS},
  year         = {2021},
  bibsource    = {dblp computer science bibliography, https://dblp.org}
}

@inproceedings{wkeglarczyk2018kernel,
  title={Kernel density estimation and its application},
  author={W{\k{e}}glarczyk, Stanis{\l}aw},
  booktitle={ITM web of conferences},
  year={2018},
}

@article{scott2010scott,
  title={Scott's rule},
  author={Scott, David W},
  journal={Wiley Interdisciplinary Reviews: Computational Statistics},
  year={2010},
  publisher={Wiley Online Library}
}
% \bibliography{mybibliography}
%
% \begin{thebibliography}{8}
% \bibitem{ref_article1}
% Author, F.: Article title. Journal \textbf{2}(5), 99--110 (2016)

% \bibitem{ref_lncs1}
% Author, F., Author, S.: Title of a proceedings paper. In: Editor,
% F., Editor, S. (eds.) CONFERENCE 2016, LNCS, vol. 9999, pp. 1--13.
% Springer, Heidelberg (2016). \doi{10.10007/1234567890}

% \bibitem{ref_book1}
% Author, F., Author, S., Author, T.: Book title. 2nd edn. Publisher,
% Location (1999)

% \bibitem{ref_proc1}
% Author, A.-B.: Contribution title. In: 9th International Proceedings
% on Proceedings, pp. 1--2. Publisher, Location (2010)

% \bibitem{ref_url1}
% LNCS Homepage, \url{http://www.springer.com/lncs}, last accessed 2023/10/25
% \end{thebibliography}
\end{document}